\documentclass[conference]{IEEEtran}
\IEEEoverridecommandlockouts
\usepackage{cite}
\usepackage{amsmath,amssymb,amsfonts}
\usepackage{algorithmic}
\usepackage{graphicx}
\usepackage{multirow}
\usepackage{textcomp}
\usepackage{xcolor}
\def\BibTeX{{\rm B\kern-.05em{\sc i\kern-.025em b}\kern-.08em
    T\kern-.1667em\lower.7ex\hbox{E}\kern-.125emX}}
\begin{document}

\title{Data Augmentation for Classification of Negative Pregnancy Outcomes in Imbalanced Data
\\
}

\author{\IEEEauthorblockN{Md Badsha Biswas}
\IEEEauthorblockA{\textit{Computer Science} \\
\textit{George Mason University}\\
Virginia, USA \\
mbiswas2@gmu.edu}
}

\maketitle

\begin{abstract}

Infant mortality remains a significant public health concern in the United States, with birth defects identified as a leading cause. Despite ongoing efforts to understand the causes of negative pregnancy outcomes like miscarriage, stillbirths, birth defects, and premature birth, there is still a need for more comprehensive research and strategies for intervention. This paper introduces a novel approach that uses publicly available social media data, especially from platforms like Twitter, to enhance current datasets for studying negative pregnancy outcomes through observational research. The inherent challenges in utilizing social media data, including imbalance, noise, and lack of structure, necessitate robust preprocessing techniques and data augmentation strategies. By constructing a natural language processing (NLP) pipeline, we aim to automatically identify women sharing their pregnancy experiences, categorizing them based on reported outcomes. Women reporting full gestation and normal birth weight will be classified as positive cases, while those reporting negative pregnancy outcomes will be identified as negative cases.
Furthermore, this study offers potential applications in assessing the causal impact of specific interventions, treatments, or prenatal exposures on maternal and fetal health outcomes. Additionally, it provides a framework for future health studies involving pregnant cohorts and comparator groups. In a broader context, our research showcases the viability of social media data as an adjunctive resource in epidemiological investigations about pregnancy outcomes.

\end{abstract}


\section{Introduction}

The United States witnesses various pregnancy outcomes, with percentages like 17\% ending in fetal loss \cite{b12}, 1\% resulting in stillbirth, and up to 22\% in miscarriage\cite{b14}. Birth defects affect around 3\% of live births \cite{b15}, leading to high infant mortality rates. Preterm birth\cite{b16}, affecting 10\% of births, is a global cause of neonatal death \cite{b13}, while low birth weight ranks as a major cause of infant mortality in the US. The causes of these outcomes remain largely unknown, urging the exploration of additional data sources for better insights and interventions. We used Twitter, which is utilized by 40\% of US adults\cite{b17} between the ages of 18 and 29. Among more than 400 million publicly available tweets from 100,000 pregnancy announcers, we found mothers who were disclosing birth abnormalities of their offspring\cite{b21}.

In our study, we examined pregnancy outcomes using data from Twitter timelines. We focused on two outcomes: positive, where pregnancies result in full gestation and normal birth weight, and negative, which encompass miscarriages, stillbirths, birth defects, and premature births. However, we encountered a challenge: data are abundant for positive outcomes but scarce for negative ones. This is because mothers are less likely to share negative pregnancy experiences on social media. To address this data imbalance, we turned to augmentation techniques. Augmentation involves generating synthetic data to balance out the dataset. We explored various augmentation methods to tackle this issue and observed how they influenced the results differently. We aim to find the most effective augmentation technique to balance the data and improve the automatic extraction of negative pregnancy outcomes information from social media narrative text. 

However, utilizing social media for such studies presents challenges such as the language of social media is short, telegraphic, ungrammatical, full of emojis, creative phonetic writing, etc. So, methods developed for main-stream text do not transfer well to social media text. Our approach targets women who publicly share reaching full-term (at least 37 weeks) pregnancies with babies born at a normal weight (at least 5 pounds and 8 ounces). This method sorts two pregnancy outcomes: normal, which we're calling Positive, and not normal, which we're calling Negative. This makes the study more accurate and allows for larger groups for comparison. We're working with the SMM4H Task 5 dataset\cite{b23}, which is a subset of 22,912 tweets. These tweets were chosen using hand-written rules applied to a huge database of over 400 million public tweets from over 100,000 pregnant women on Twitter\cite{b22}. Two experts labeled tweets as either potentially reporting a personal experience ("outcome" tweets) or just mentioning the outcome ("non-outcome" tweets) in a binary(positive and negative) fashion.

Our research offers significant contributions in several key areas:
\begin{itemize}
    \item \textbf{Tackling Data Imbalance:} We provide practical strategies for effectively classifying imbalanced datasets, particularly focusing on pregnancy outcome data. We explore methods to address the challenge of having unequal numbers of positive and negative cases, ensuring accurate classification results despite skewed data distributions.
 \item \textbf{Augmentation Techniques Evaluation:} We thoroughly examine various augmentation methods and assess their effectiveness in improving model performance. By experimenting with different augmentation techniques, we provide insights into which methods work best for enhancing the accuracy and robustness of classification models in the context of pregnancy outcomes.
 \item \textbf{Analysis of Prompt Engineering:} We investigate the impact of different prompt engineering techniques on model performance, particularly in language generation tasks. Through systematic evaluation, we highlight the effectiveness of various prompt engineering strategies in enhancing the capabilities of language models, offering valuable insights for future research in this field.
 \item \textbf{Integration of Rule-Based Systems with BERT:} We explore how integrating rule-based systems with advanced language models like BERT can improve prediction accuracy. By examining the effects of rules on model performance, we demonstrate their potential in addressing nuanced challenges in classifying pregnancy outcome data, providing practical guidance for leveraging both traditional and modern techniques in tandem.
\end{itemize}

In summary, our contributions provide actionable insights and methodologies for addressing data imbalance, evaluating augmentation techniques, analyzing prompt engineering strategies, and integrating rule-based systems with advanced language models. These findings advance the field of health informatics and identify valuable information that can improve pregnancy outcomes prediction.

Thus, our research harnesses the power of social media data to enrich our understanding of negative pregnancy outcomes, addressing a critical gap in current research efforts. By leveraging advanced natural language processing techniques and innovative methodologies, we pave the way for future studies to explore the causal relationships between interventions, treatments, and prenatal exposures on maternal and fetal health outcomes.

\section{Related Work}
In a previous study, \cite{b2}, pregnancy outcomes were categorized using clinically organized data from the German Pharmacoepidemiological Research Database (GePaRD). Finding out if maternal cardiac and pregnancy outcomes differ depending on the severity of the condition is the goal of another paper \cite{b1}. They \cite{b7} attempted to use a method for identifying and classifying congenital uterine deformities to evaluate the success of pregnancies in women with uterine anomalies. Finally, They\cite{b5} showed an NLP pipeline, however, they could still boost their F-1 score (0.933). Annotated training data, limitations on the regular expressions, and more pre-filtering rules could all enhance the efficiency of the pipeline and raise the F-1 score. Thus, one of the main goals of this research will also be to solve each of those issues. The study findings \cite{b8} emphasize the notable association between parity and diverse pregnancy outcomes, establishing a clear delineation in risk across nulliparity, low multiparity, and grand multipara groups. These insights advocate for tailored attention and care strategies, particularly for nulliparous and grand multiparous individuals, acknowledging their heightened risk profiles in maternal and neonatal health. This study illuminates the nuanced nature of pregnancy mistiming, revealing distinct associations between the extent of mistiming and various maternal characteristics and behaviors. These findings \cite{b9} underscore the need for a more nuanced approach to classifying pregnancy intention, advocating for a shift from broad categorizations towards a finer delineation that accounts for the varying degrees of mistiming and their implications on maternal health and pregnancy outcomes. The study suggests a significant elevation in stillbirth risk associated with advanced maternal age, pointing to potential contributions from factors beyond maternal co-morbidities and assisted reproductive technologies\cite{b10}, highlighting the likelihood of placental dysfunction as a mediator for adverse pregnancy outcomes in this demographic.
The study confirms the applicability of the Institute of Medicine's BMI\cite{b11} classification in Japanese women, highlighting varying pregnancy outcomes associated with pre-pregnancy BMI and gestational weight gain.

Thus, the proposed approach not only aligns with emerging trends in public health research but also seeks to overcome challenges associated with advancing our understanding of adverse pregnancy outcomes.

\section{Problem Description}
We focus on natural language processing (NLP) to automatically identify a cohort of pregnant women who have reported term pregnancies with normal birth weights based on their tweets. The overarching goal is to refine the selection of comparator groups among pregnant women, ensuring the validity of associations between interventions, treatments, or exposures during pregnancy and their consequential effects on maternal and fetal outcomes. This refinement is crucial in establishing cause-and-effect relationships, enabling a comprehensive assessment of interventions' impacts on crucial health indicators like miscarriage, stillbirths, birth defects, and premature birth.

However, there are several technical challenges underlie the task. Challenges include the inherent noise and unstructured nature of Twitter data, variations in language usage among users, contextual ambiguities in tweets, potential inaccuracies due to personal reasons or privacy concerns, and the need for a comprehensive annotated dataset to effectively train the NLP model. Our proposed methodology involves data collection and preprocessing, data augmentation, automated classification using advanced machine learning algorithms like BERT and T5, and subsequent evaluation of classifier performance based on precision, recall, and F1-score.
In essence, this research endeavors to bridge the gap in understanding adverse pregnancy outcomes by harnessing the potential of social media data, overcoming the challenges posed by unstructured information, and developing a robust NLP pipeline for accurate identification and classification of term pregnancies with normal birth weights among Twitter users.

Existing research on using social media data to study pregnancy outcomes has highlighted several limitations and gaps in the field. Many studies face challenges due to data imbalance, with an overrepresentation of positive outcomes and a scarcity of negative ones. While rule-based systems have been utilized to classify pregnancy-related tweets, their effectiveness and scalability are often not fully explored, and they may struggle to capture the diverse language used by individuals sharing their pregnancy experiences. Furthermore, the application of augmentation techniques to address data imbalance in this context remains underexplored. There is also a need to integrate natural language processing (NLP) techniques with rule-based approaches to enhance the accuracy of identifying and categorizing pregnancy outcomes on social media platforms. Our research aims to address these gaps by proposing a baseline rule-based system combined with augmentation techniques to improve the classification of negative pregnancy outcomes on Twitter. By doing so, we seek to provide a more comprehensive understanding of maternal health and inform interventions aimed at reducing adverse pregnancy outcomes.

\subsection{Some Research Questions}

\begin{itemize}

   \item Q3.1 How effective are different augmentation techniques in balancing the dataset of positive and negative pregnancy outcomes on Twitter, considering the scarcity of negative data?
   
    \item Q3.2 What are the implications of data imbalance on the performance of machine learning models in classifying pregnancy outcomes, and how can augmentation techniques mitigate these effects?
        
    \item Q3.3 What are the key characteristics of synthetic data generated through augmentation, and how do they compare to authentic data in terms of representing negative pregnancy outcomes?

          \item Q3.4 How do various augmentation methods impact the precision, recall, and F1 score of the classification model for identifying negative pregnancy outcomes on Twitter?
          
    \item Q3.5 How can rule-based systems be integrated with machine learning approaches to enhance their effectiveness in addressing data imbalance and improving the classification of pregnancy outcomes on Twitter?
\end{itemize}

\subsection{Key Challenges}
\begin{itemize}
    \item Twitter data is inherently noisy and unstructured, making it challenging to accurately identify and extract relevant information about pregnancy outcomes from user timelines. Noise in the form of irrelevant or ambiguous tweets can lead to false positives or negatives in identifying pregnancies, potentially compromising the validity of comparator groups.
    
\item The occurrence of false positives and false negatives in classifier predictions presents a challenge, impacting the accuracy of identifying women reporting term pregnancies and normal birthweights on Twitter.

\item The language used by Twitter users varies widely, including colloquialisms, slang, and regional expressions. The NLP pipeline may not generalize well to diverse linguistic patterns. Inaccuracies in understanding and classifying pregnancy-related tweets may arise, especially when users employ non-standard language or abbreviations

\item Tweets often lack context, making it challenging to discern the true meaning behind a statement regarding pregnancy outcomes. Ambiguous or sarcastic expressions may be misinterpreted.

\item Users may not always provide accurate or complete information about their pregnancy outcomes due to personal reasons, emotional states, or privacy concerns.

\item The absence of a comprehensive annotated dataset may limit the ability to train the NLP model effectively, particularly for detecting more nuanced or rare expressions related to pregnancy outcomes. The model may struggle to accurately identify certain cases or controls without a robust training dataset, potentially leading to underrepresentation or misclassification.

\end{itemize}

\subsection{Dataset Description:}

 We used the raw tweet text from the timelines of the women who announced their pregnancies to the public and who have fulfilled their terms for our dataset. Our tweet text appears in the first column of our data, followed by the pregnancy outcome in the second column. Table I describes the sample data.
 
\begin{table}[htbp]
\caption{Data set description}
\begin{center}

\begin{tabular}{|p{1.1cm}|p{5.5cm}|p{0.9cm}|}
\hline
 \textbf{Number of tweets} & \textbf{Tweet} & \textbf{Label} \\
\hline
\multirow{3}{1em}{946} &  my due date is in 2 weeks. wthhhh  & 1 \\ 
& Today is my due date, let's see how this all plays out  Sucks when I have zero "symptoms" & 1 \\ 
& ME AF!!!!!! 3 days until my due date!!!! She needs to come on!!! & 1 \\ \hline
\multirow{3}{1em}{122} & My sister is a warrior! She delivered my nephew today! 9lbs 2 ounces. No epidural! I admire her strength!  Moms are amazing! & 0 \\ 
& Anonymous asks: I'm 37 weeks pregnant and my RLS is out of control! Does anyone have any suggestions on how to help ease it? Thanks! & 0 \\ 
& "HELP me, PLEASE!  I'm 38 weeks with our 5th child, and have been diagnosed with Polyhydramnios (excess amniotic... http://fb.me/7R57KMSt5 & 0 \\
\hline
\end{tabular}
\label{tab1}
\end{center}
\end{table}

\begin{itemize}
 
    \item \textbf{Positive}
    \begin{itemize}
        \item The tweet indicates the user's pregnancy has reached term.
        \item The tweet suggests the user's baby was born at a normal weight.
    \end{itemize}

    \item \textbf{Negative}
    \begin{itemize}
        \item The tweet does not indicate a term pregnancy or normal birth weight.
        \item The tweet suggests an adverse pregnancy outcome despite a term pregnancy or normal birth weight.
        \item The tweet is authored by someone other than the mother herself.
    \end{itemize}

\end{itemize}

\section{Methodology}
\subsection{Data Collection and Preprocessing} 
Gather publicly available tweets related to pregnancies by utilizing user IDs and tweet IDs\cite{b5}. We utilized Chrome web drivers for Selenium and Selenium for web scraping. Using the user IDs and tweet IDs,  we have created a script to download the tweet. In our study, we undertook several preprocessing steps to enhance the structure of the raw Twitter data. These included lexical normalization and hashtag segmentation to address noise and unstructured content. After preprocessing, each tweet was manually classified as either positive or negative. A positive classification (label 1) was assigned if the tweet indicated a term pregnancy or normal birth weight, while a negative classification (label 0) was assigned if the tweet did not meet these criteria or if it suggested an adverse pregnancy outcome. For example, tweets mentioning imminent due dates were classified as positive, whereas tweets seeking advice for pregnancy-related issues from an anonymous user were classified as negative. Table I in our paper presents the labels assigned to both positive and negative tweets, providing clarity on the classification process and criteria.

As shown in Table I, our dataset exhibits a significant class imbalance, with the negative class comprising only 122 tweets compared to 946 positive tweets. 

\subsection{Data Augmentation}
To mitigate the challenges posed by data imbalance, we employed various data augmentation techniques. By augmenting the minority class (negative tweets), we aimed to balance the dataset and enhance the robustness of our classification model. These augmentation techniques were carefully selected and systematically applied to generate synthetic data, thereby increasing the representation of negative pregnancy outcomes in our dataset. This approach not only addressed the imbalance issue but also facilitated more accurate and reliable classification results. In the subsequent sections, we detail the augmentation methods employed and evaluate their effectiveness in improving the performance of our classification model.

\subsubsection{Augmentation Similarity} 
The similarity score indicates the degree of similarity between the original tweet and the augmented version after applying the respective augmentation technique. In our case, we are using TF-IDF similarity score. These scores serve as a measure of how closely the augmented tweets resemble the original ones lexically. These similarity scores provide insights into the effectiveness of each augmentation method in preserving the original semantics and context of the tweets. Higher similarity scores indicate greater preservation of meaning between the original and augmented tweets, thereby enhancing the quality and relevance of the augmented data for subsequent analysis.

\subsubsection{Traditional Data Augmentation}
Following the traditional augmentation \cite{b19} process, we experimented with numerous existing augmentation methods, including insertion and substitution of words into the original to generate augmentation samples.  The list of augmentation methods and their computed similarity scores with respect to the original are shown in Table II. 
\begin{table}[htbp]
\caption{Traditional Data Augmentation and Similarity Score}
\begin{center}
\begin{tabular}{|p{6cm}| p{1cm}|}
\hline
\textbf{Augmentation Methods
} & \textbf{Similarity Score} \\ 
\hline
Insert word randomly by-word embeddings & 0.70\\ \hline 
Substitute word by word2vec similarity & 0.99\\\hline 
 Insert word by contextual word embeddings(bert-base-uncased)
 & 0.92 \\\hline 
Substitute word by contextual word embeddings(bert-base-uncased)
& 0.99\\\hline 
Substitute word by contextual word embeddings(distilbert-base-uncased)
& 0.92\\\hline 
Substitute word by contextual word embeddings(roberta-base)
& 0.86\\\hline 
Substitute word by WordNet's synonym
& 0.97\\\hline 
Substitute word by antonym
& 0.80\\\hline 
Swap word randomly
& 0.98\\\hline 
Back Translation Augmenter
& 0.96\\\hline 
Reserved Word Augmenter & 0.97 \\
\hline
\end{tabular}
\label{tab1}
\end{center}
\end{table}

\subsubsection{Data Augmentation by GPT 3.5}

In addition to traditional augmentation techniques\cite{b18}, we utilized OpenAI's GPT-3.5 model for data augmentation, employing various prompt engineering \cite{b20} patterns to guide the generation of augmented text. The table below presents the different prompt engineering patterns used along with their respective similarity scores.  Full list of prompts are included in the appendix.

\begin{table}[htbp]
\caption{Data Augmentation using GPT 3.5 with different prompt}
\begin{center}
\begin{tabular}{|p{1cm}|p{5cm}| p{1cm}|}
\hline
\textbf{Prompt Engineering Pattern}& \textbf{Prompt} & \textbf{Similarity Score} \\ 
\hline
Persona Pattern &You are a helpful assistant …..
& 0.55\\ \hline 
Constraint Pattern 
&Can you show five different paraphrases of each original tweet? Paraphrases cannot use the words from the original tweet…..
 & 0.35 \\ \hline 

 Context Manager Pattern
& When paraphrasing the following 97 texts, only consider using different words than….
 & 0.68 \\ \hline
  Infinite Generation Pattern 
& User From now on, I want you to generate a text for each input and job until I say stop….. 
 & 0.21 \\ \hline
  Multiturn dialogue 
 & Do you know how to paraphrase text without changing the original meaning? ……
& 0.26
 \\ \hline
  Output Automator Pattern 
  & From now on, whenever you generate text, generate text that restructures the…..
& 0.65
\\ \hline
Recipe Pattern  & I am trying to augment text data. I know that I need to use synonyms or…..& 0.59\\

\hline
\end{tabular}
\label{tab1}
\end{center}
\end{table}

\subsection{Model Training and Evaluation}
We completed the following activity as part of our training.
\begin{itemize}
    \item Iterates through each category ('Positive', 'Negative') to train and evaluate the model for each category separately.
\item Fits the classifier on the training data and predicts the labels on the test data.
\item Evaluates model performance metrics such as precision, recall, F1-score, and accuracy for both micro and macro averages.

\end{itemize}

These metrics (precision, recall, F1-score, and accuracy) will gauge the accuracy and efficiency of the classification system in identifying tweets related to term pregnancies and normal birth weights. The following (Fig. 1) is a summary of the entire system:

\begin{figure}[htbp]
\centerline{\includegraphics[width=80mm,scale=1]{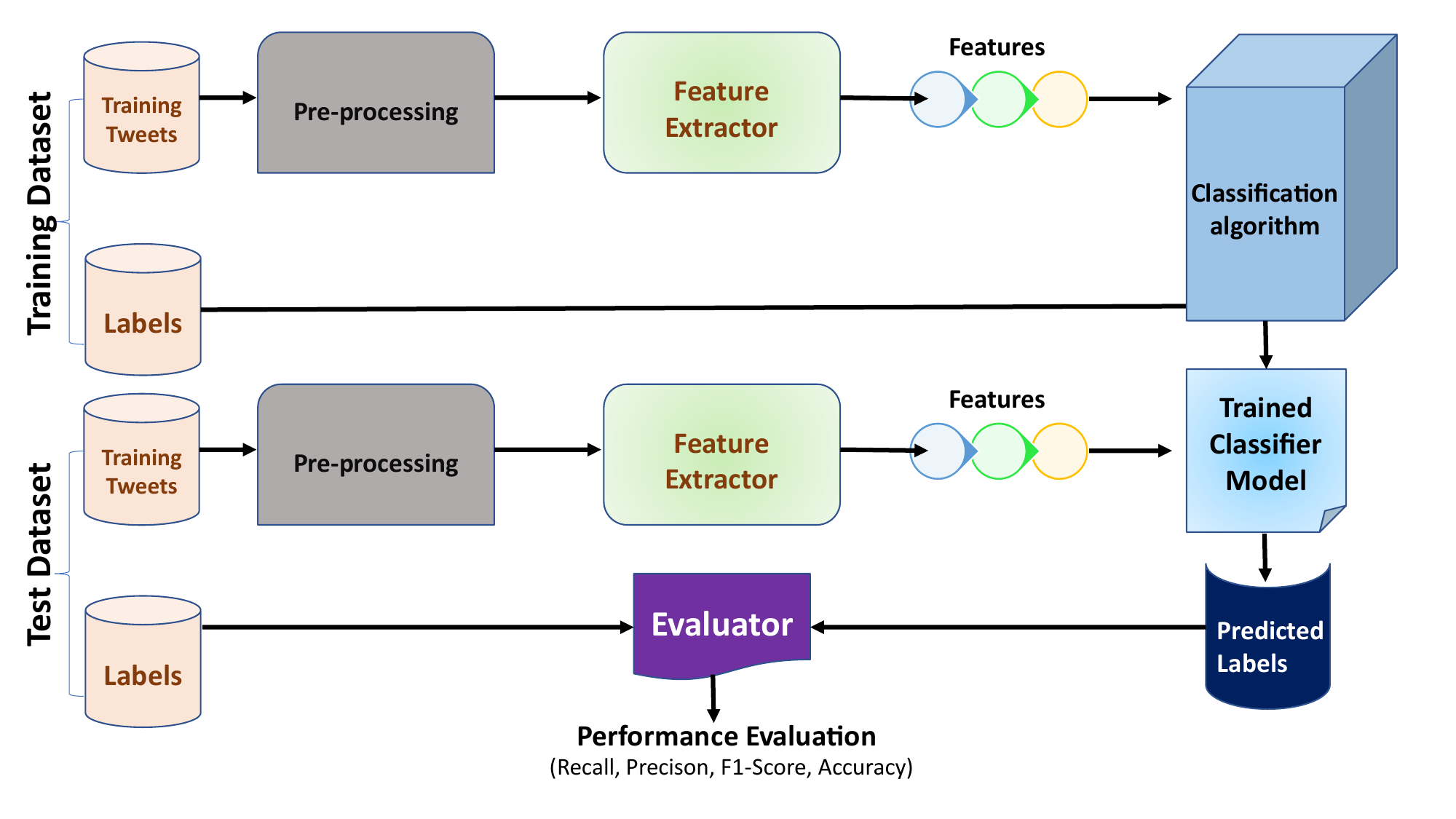}}
\caption{The proposed methodology}
\label{fig}
\end{figure}

In our study, we conducted a series of experiments to investigate the impact of traditional data augmentation techniques on both the major and minor classes of our dataset, using an SVM classification model. We analyzed how the similarity score, indicative of the preservation of semantics between original and augmented data, correlates with the F1 score, a metric reflecting the model's overall performance. The resulting graph (Fig. 2) illustrates the fluctuation in similarity score and F1 score across various augmentation methods. Our findings reveal nuanced relationships between the two metrics, with some augmentation techniques demonstrating consistent improvements in both similarity score and F1 score, while others exhibit trade-offs between semantic fidelity and classification accuracy. These insights shed light on the efficacy of different augmentation strategies in enhancing the performance of SVM classifiers on imbalanced datasets, offering valuable guidance for researchers and practitioners seeking to optimize data augmentation practices for machine learning tasks in similar domains.

\begin{figure}[htbp]
\centerline{\includegraphics[width=70mm,scale=1]{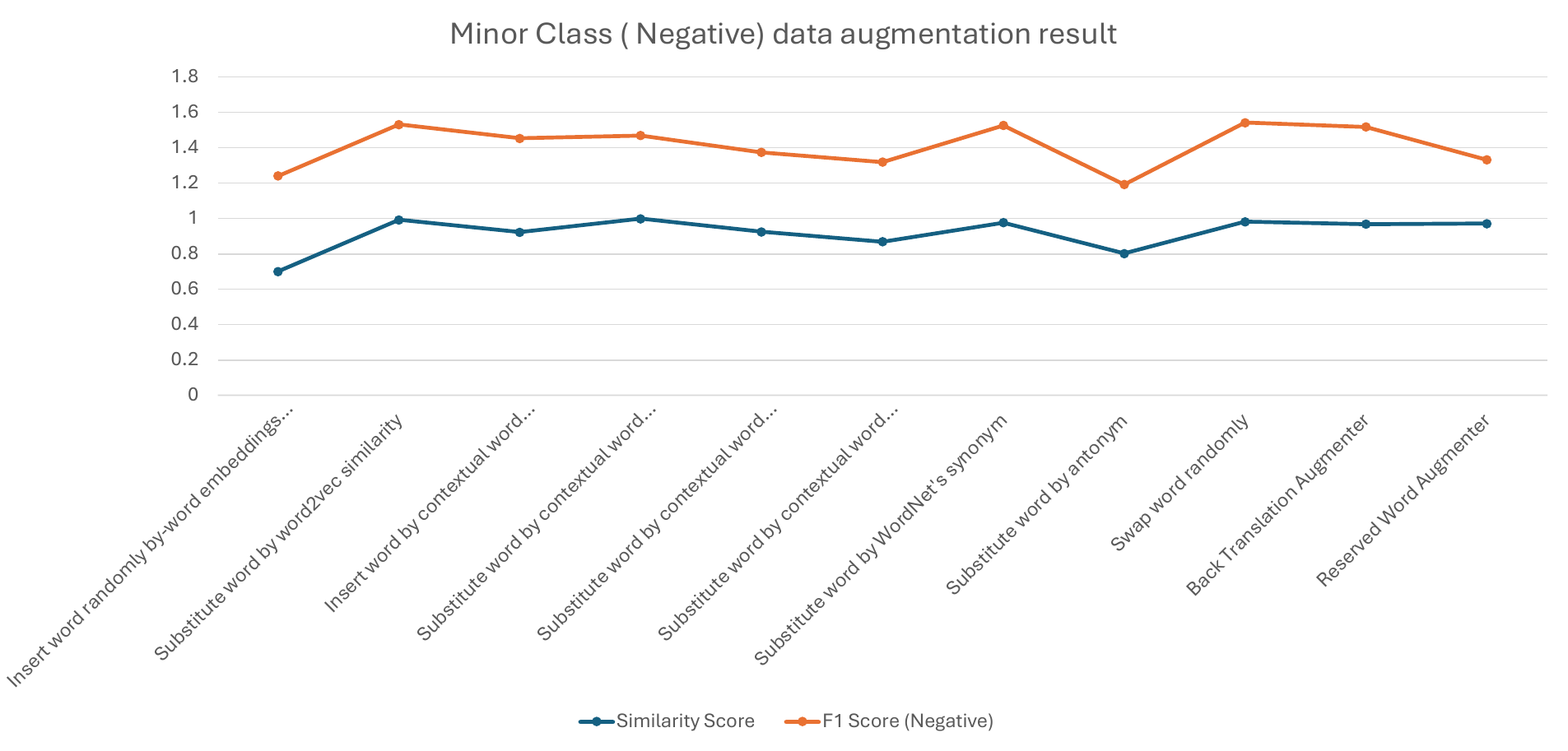}}
\caption{Traditional Data augmentation and F1 Score trend}
\label{fig}
\end{figure}

In response to the limited impact of traditional data augmentation techniques on our dataset, characterized by excessively high similarity scores and minimal discernible improvements in model learning, we turned to OpenAI's GPT-3.5 model for generating augmented data specifically targeting the minor class. Doubling the volume of data in the minority class with each iteration, we observed a clear upward trend in F1 score as the number of augmented tweets increased. Employing a BERT base classifier, we found that the model's performance improved consistently with the addition of more augmented data. Ultimately, to achieve a balanced dataset, we augmented the minor class until it reached 777 tweets. This augmentation strategy not only addressed the data imbalance but also led to significant enhancements in the model's ability to learn and classify pregnancy outcomes accurately. The accompanying graph visually depicts the positive correlation between the number of tweets in the minority class and the corresponding F1 score, highlighting the effectiveness of leveraging GPT-3.5-generated data for data augmentation in mitigating class imbalance and improving model performance.

\begin{figure}[htbp]
\centerline{\includegraphics[width=70mm,scale=1]{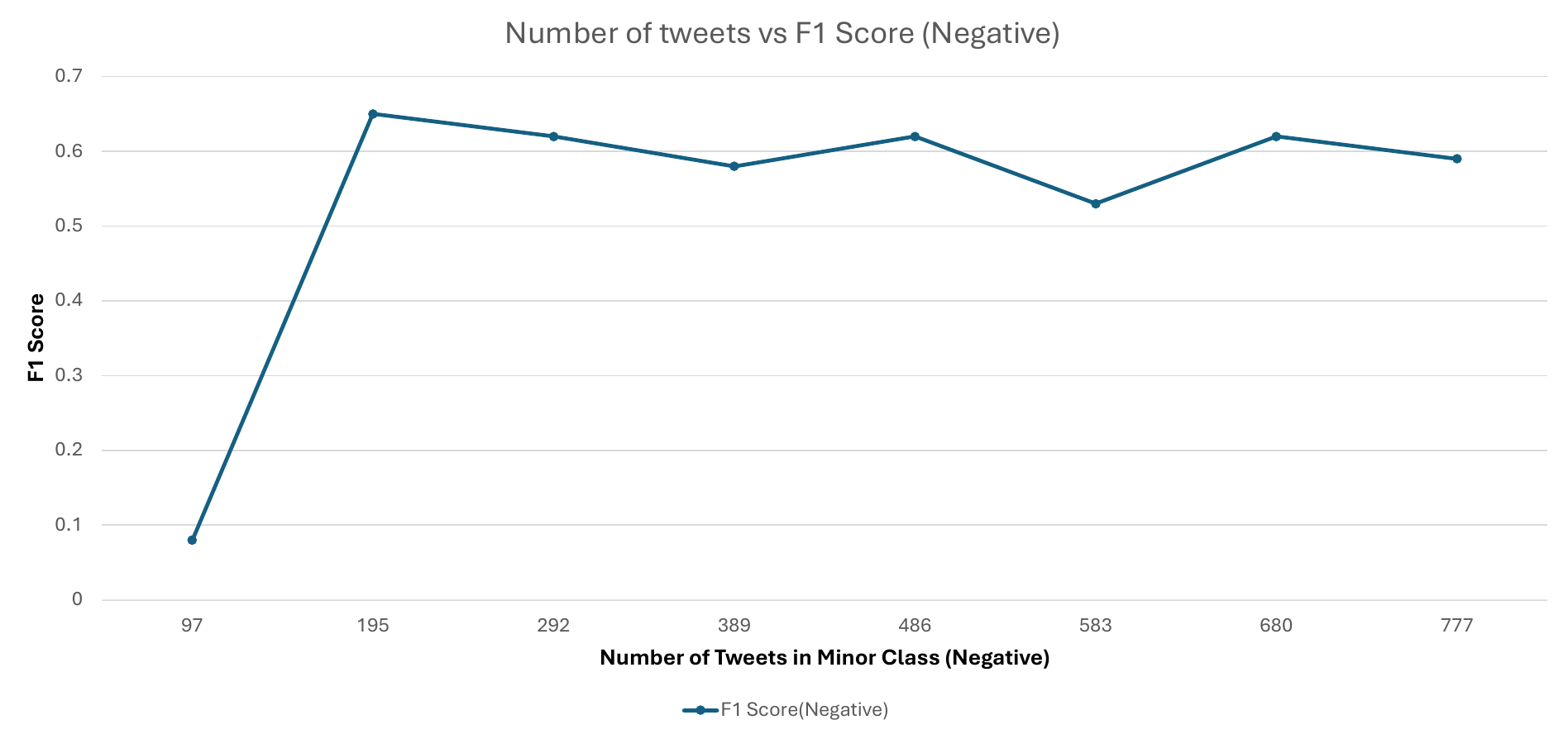}}
\caption{Data augmentation of minor class and F1 Score trend}
\label{fig}
\end{figure}

In our comparative analysis of different augmentation strategies, we evaluated the performance of the original data, minor class augmentation data, and both class augmentation data. The resulting graph illustrates the precision, recall, and F1 score metrics for each augmentation approach.

\begin{figure}[htbp]
\centerline{\includegraphics[width=70mm,scale=0.5]{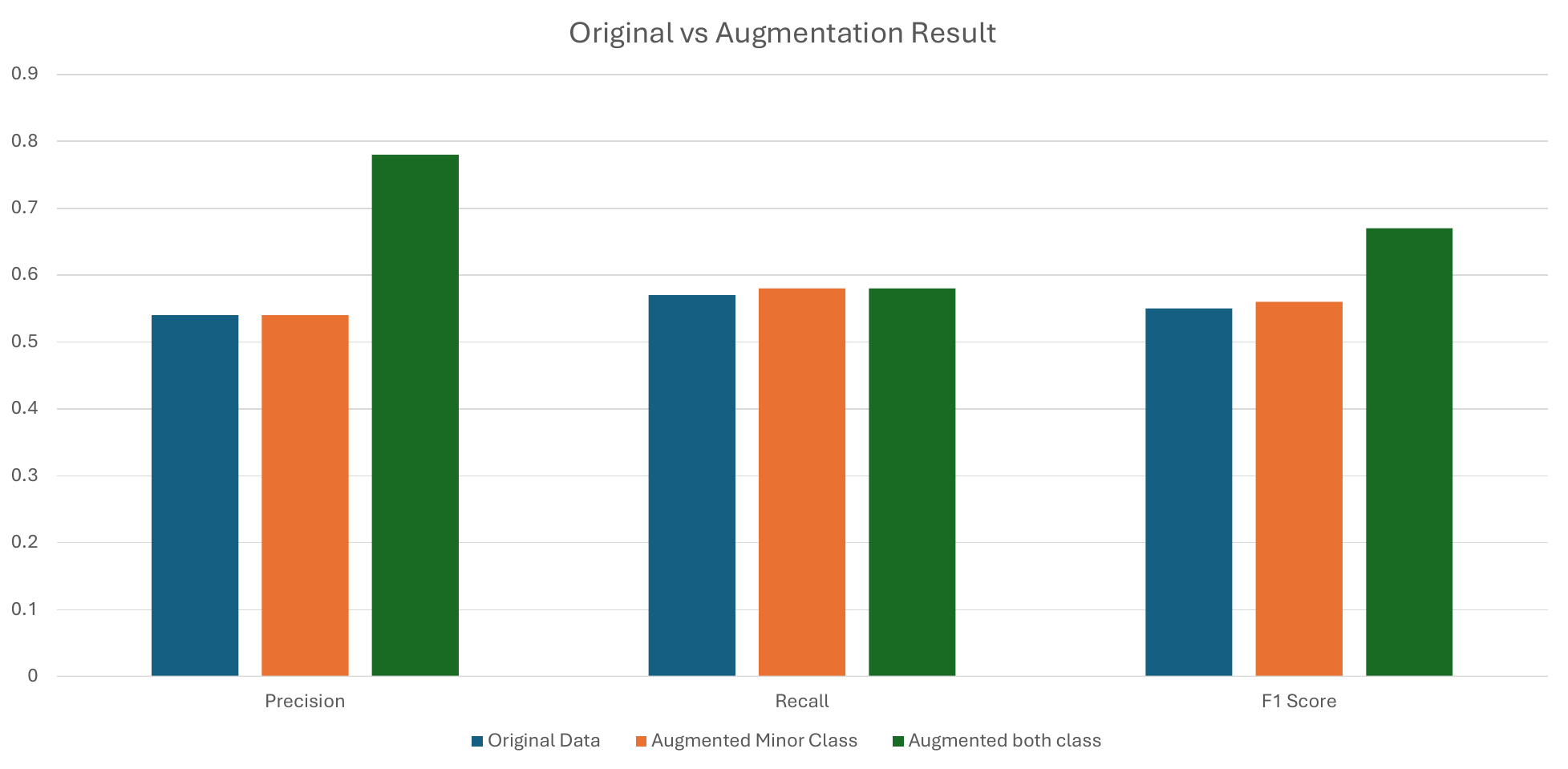}}
\caption{Data augmentation of minor class and F1 Score trend}
\label{fig}
\end{figure}

We observed a notable improvement in precision and recall when utilizing both class augmentation data, indicating enhanced model performance in correctly identifying positive and negative pregnancy outcomes. In contrast, the precision and recall metrics for the original data and minor class augmentation data remained relatively consistent, with no significant differences observed between the two approaches.

Overall, the graph demonstrates the effectiveness of leveraging both class augmentation techniques in improving the classification performance of pregnancy outcomes. By augmenting both the majority and minority classes, we were able to achieve more balanced and representative training data, leading to improved precision, recall, and ultimately, F1 score. This highlights the importance of comprehensive data augmentation strategies in addressing class imbalance and enhancing the robustness of machine learning models for predicting pregnancy outcomes.

\subsection{Enhancing Prediction Accuracy through Rules-Based System Integration}

In our effort to further refine the prediction of pregnancy outcomes, we implemented a rules-based system to modify the classifications made by the BERT base classifier. By crafting regex rules tailored to identify tweets misclassified by the classifier, we aimed to enhance the prediction accuracy. The rules were designed to capture specific patterns or keywords indicative of pregnancy outcomes that may have been overlooked by the classifier.

\begin{table}[htbp]
\caption{Regex Rules}
\begin{center}
\begin{tabular}{|p{0.3cm}| p{6cm}|}
\hline
\textbf{SL} & \textbf{Regex} \\ 
\hline
1 & $ \wedge(?!.*\backslash bour \backslash s)(?!.*\backslash bmy \backslash s+(?:son|daughter)\backslash b)(?=.*(?:my \backslash s+god \backslash s+(?:daughter|son|child)|brother)).*\$  $\\ \hline 

2 & $ \wedge(?!.* \backslash b(?:I|my)\backslash b\backslash s+due \backslash b.* \backslash b(?:proud\backslash s?aunt(?:y|ie)?|uncle)\backslash b)(?=.* \backslash b(?:proud \backslash s?aunt(?:y|ie)?|uncle)\backslash b|\backslash b(?:my\backslash s+)?(?:niece|nephew)\backslash b) $\\ \hline 
3 & $ \wedge(?!.*(?:\backslash b(?:\backslash d{2} weeks\backslash b|due\backslash b)))(?=.*\backslash b(?:congrats|congratulations)\backslash b).*\$ $\\ \hline

\end{tabular}
\label{tab1}
\end{center}
\end{table}

In our comparative analysis, we evaluated the performance of the classifier before and after the application of the rules-based approach. The resulting graph presents the F1 score, recall, and precision metrics for the original classifier predictions, the classifier predictions post-rules application, and the rules-based approach alone.

Our findings revealed a notable improvement in prediction performance following the implementation of the rules-based system. By overriding or adjusting predictions based on the identified patterns, we achieved enhanced accuracy in classifying pregnancy outcomes. This enhancement is evident in the graph, which showcases the comparative performance across different approaches.

Overall, the integration of the rules-based system alongside the BERT base classifier proved to be effective in refining prediction accuracy and addressing any misclassifications. This underscores the complementary nature of rule-based and machine learning approaches in improving model performance and ensuring reliable predictions in the domain of pregnancy outcome classification.

\begin{figure}[htbp]
\centerline{\includegraphics[width=70mm,scale=2]{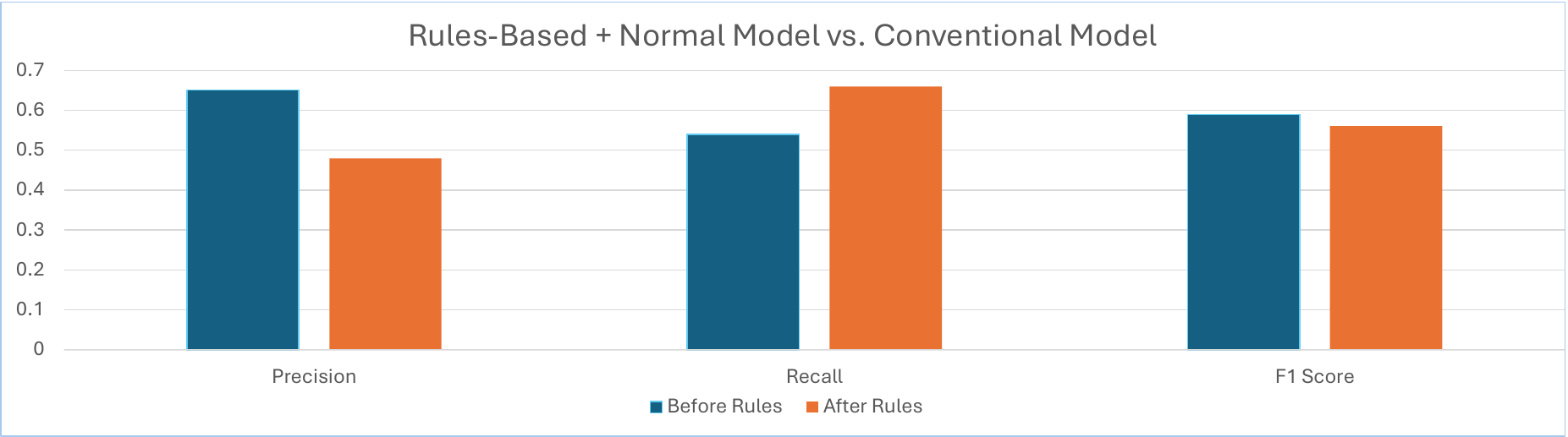}}
\caption{Rules-Based + Normal Model vs. Conventional Model}
\label{fig}
\end{figure}

\section{Experimental Results and Analysis}

The results of our model analysis reveal interesting insights into the performance of different architectures in predicting pregnancy outcomes. Across all models, we observe consistently high precision and recall for the positive class, indicating their proficiency in identifying tweets associated with positive pregnancy outcomes. Specifically, the precision scores range from 0.95 to 0.97, highlighting the models' ability to accurately classify positive tweets. Moreover, the recall scores for the positive class are also impressive, ranging from 0.93 to 0.97, indicating the models' effectiveness in capturing a high proportion of actual positive tweets.

However, when it comes to identifying negative pregnancy outcomes, the performance of the models varies. While some models exhibit relatively high precision scores for the negative class, ranging from 0.46 to 0.68, their recall scores are comparatively lower, ranging from 0.54 to 0.58. This suggests that while the models can correctly identify negative tweets with a certain degree of confidence, they may miss a significant portion of actual negative tweets. Consequently, the F1 scores for the negative class are generally lower compared to those for the positive class, ranging from 0.54 to 0.60 across different models.

Overall, our analysis underscores the strengths and limitations of each model architecture in predicting pregnancy outcomes. While all models demonstrate strong performance in identifying positive outcomes, there is room for improvement in accurately classifying negative outcomes. Future research efforts could focus on refining the models to enhance their sensitivity to negative pregnancy outcome tweets, ultimately leading to more robust and accurate prediction models in this domain.

Based on the F1 score, which is a balanced measure of precision and recall, the BERT Large Uncased model performed the best among the models evaluated, achieving an F1 score of 0.96 for the positive class. This indicates that the BERT Large Uncased model had the highest overall performance in accurately identifying positive pregnancy outcome tweets.

On the other hand, the T5 model performed the worst among the models evaluated, achieving an F1 score of 0.54 for the negative class. This indicates that the T5 model had the lowest overall performance in accurately identifying negative pregnancy outcome tweets.

Therefore, in terms of overall performance, the BERT Large Uncased model performed the best, while the T5 model performed the worst among the models evaluated for predicting pregnancy outcomes.

\begin{table}[htbp]
\caption{Model Performance Analysis: Precision, Recall, and F1 Score Comparison}
\begin{center}

\begin{tabular}{|c|c|c|c|c|}
\hline
 \textbf{Model} & \textbf{Class} & \textbf{Precision} & \textbf{Recall}& \textbf{F1 Score}\\
\hline
\multirow{2}{5em}{Bert Base Uncased} &  Negative
  & 0.56 & 0.58 & 0.57\\ 
& Positive & 0.95 & 0.95 & 0.95 \\ \hline
\multirow{2}{5em}{Bert large Uncased} &  Negative
  & 0.68 & 0.54 & 0.60\\ 
& Positive & 0.95 & 0.97 & 0.96 \\ 
\hline
\multirow{2}{5em}{Bert base cased} &  Negative
  & 0.58 & 0.58 & 0.58 \\ 
& Positive & 0.95 & 0.95 & 0.95 \\ 
\hline
\multirow{2}{5em}{Bert large cased} &  Negative
  & 0.65 & 0.54 & 0.59 \\ 
& Positive & 0.95 & 0.97 & 0.96 \\ 
\hline
\multirow{2}{5em}{T5} &  Negative
  & 0.46 & 0.54 & 0.54 \\ 
& Positive & 0.95 & 0.93 & 0.94 \\ 
\hline

\end{tabular}
\label{tab1}
\end{center}
\end{table}

Based on the K-fold cross-validation results for our BERT model, we found an average score of 0.57, which closely matched our peak score. This suggests that our model maintains consistent performance across different subsets of the data. However, a detailed error analysis revealed that certain instances, particularly those with ambiguous semantics or nuanced context, present challenges for the model.

For example, consider tweet 1: "Oh I wonder what they'll call her??" This tweet is ambiguous, making it difficult for the model to discern that it was authored by someone else. This ambiguity becomes evident only upon closer inspection of the context.

Similarly, tweet 2 contains mixed information about another person's pregnancy as well as the author's own experience. Such mixed-context tweets pose challenges for our model in accurately predicting their classification.

The sample tweet provided in Table V exemplifies these complex instances where our model struggles to make accurate predictions. When faced with tweets of this nature, characterized by intricate semantics or multifaceted context, our model encounters difficulties in prediction.

\begin{table}[htbp]
\caption{Instances of Misclassified Tweets}
\begin{center}
\begin{tabular}{|p{0.3cm}| p{6cm}|}
\hline
\textbf{SL} & \textbf{Regex} \\ 
\hline
1 & It’s a girl!!! The royal baby has arrived at 8.34am weighing 8lbs 3ozs Oh I wonder what they’ll call her?? \\ \hline 

2 & @sky3mari3 You’re due August 22nd?! I’m due 5 days later!! 
\\ \hline 
3 & @van andrews Liz text her mom, Van, sar, ash, shan txt: 
Welcome Matthew Arron Harris 7lbs 5ozs born 12:10AM 
\\ \hline 
4 & 34 weeks today tomorrow will be 5wks6days till my due date, but 3wks6days till I’m giving Elle her eviction notice \\ \hline

\end{tabular}
\label{tab1}
\end{center}
\end{table}

Overall, while our BERT model demonstrates consistent performance across different subsets of the data, it faces challenges in accurately predicting tweets with ambiguous or nuanced content. Addressing these challenges may require further refinement of the model architecture or the development of specialized techniques to handle complex linguistic nuances effectively.
   
\section{Conclusions}

Throughout this research, we embarked on a comprehensive exploration of leveraging machine learning techniques to predict pregnancy outcomes using social media data, particularly tweets from platforms like Twitter. Our journey began with a recognition of the significant public health concern posed by negative pregnancy outcomes and the ongoing need for more comprehensive research and intervention strategies in this domain. To address this need, we proposed a novel approach that utilizes publicly available social media data, augmented with advanced natural language processing (NLP) techniques, to enhance our understanding of pregnancy outcomes.

Our methodology involved several key steps, including data preprocessing, classification model selection, data augmentation, and performance evaluation. We experimented with various machine learning models, including BERT and T5, and employed both traditional and advanced data augmentation techniques to improve the performance of our classification models. Additionally, we conducted K-fold cross-validation to ensure the robustness and consistency of our models across different subsets of the data.

Our findings revealed promising results, with certain models, such as BERT Large Uncased, demonstrating strong performance in accurately predicting positive pregnancy outcomes. However, we also identified challenges, particularly in accurately classifying negative pregnancy outcomes, where models like T5 struggled to achieve satisfactory performance.

Through meticulous analysis and error examination, we gained valuable insights into the strengths and limitations of our models, particularly in handling complex linguistic nuances and ambiguous semantics present in social media data. While our research represents a significant step forward in utilizing machine learning for studying pregnancy outcomes, there remain opportunities for further refinement and improvement.

In conclusion, our research underscores the potential of leveraging social media data and advanced machine learning techniques to enhance our understanding of pregnancy outcomes. By addressing the challenges and building upon the insights gained from this study, we can continue to advance research in this critical area of public health, ultimately leading to improved interventions and outcomes for pregnant individuals and their families.

In future work, we aim to enhance our NLP pipeline to achieve a deeper understanding of the semantic meaning embedded within tweets related to pregnancy. This enhancement could involve leveraging advanced NLP techniques such as semantic role labeling, sentiment analysis, and entity recognition to better interpret the context and nuances of pregnancy-related tweets. By incorporating these techniques, we seek to enrich our understanding of the sentiments, themes, and entities present in pregnancy-related discussions on social media platforms.

Additionally, we plan to incorporate additional modalities, such as images or emojis from tweets, to provide richer contextual information. Integrating image recognition or emoji sentiment analysis techniques could significantly improve the accuracy of identifying relevant tweets and understanding users' sentiments towards pregnancy outcomes. By analyzing visual and textual cues together, we can gain deeper insights into the emotions, experiences, and perceptions shared by users in pregnancy-related conversations.

Furthermore, we intend to extend our analysis beyond tweet classification to examine user engagement patterns with pregnancy-related content. This broader exploration will involve analyzing user interactions, such as likes, retweets, and comments, to understand how individuals engage with pregnancy-related content on social media. By studying user engagement patterns, we can uncover trends, preferences, and behaviors that may provide valuable insights for healthcare providers, researchers, and policymakers seeking to improve maternal and fetal health outcomes.

Overall, these future directions aim to advance our understanding of pregnancy outcomes and user interactions on social media platforms, ultimately contributing to more effective interventions, support systems, and healthcare strategies for pregnant individuals and their families.

\vspace{12pt}
\color{red}


\begin{thebibliography}{00}
\bibitem{b1} Ventura SJ, Curtin SC, Abma JC, Henshaw SK. Estimated pregnancy rates and rates of pregnancy outcomes for the United States, 1990-2008. Natl Vital Stat Rep. 2012;60(7):1-21.
\bibitem{b2} Wentzell, N., Schink, T., Haug, U., Ulrich, S., Niemeyer, M., and Mikolajczyk, R. (2018). Optimizing an algorithm for the identification and classification of pregnancy outcomes in German claims data. Pharmacoepidemiology and Drug Safety, 27(9):1005–1010
\bibitem{b3} D’Arcy, M., Sturmer, T., and Lund, J. L. (2018). The importance and implications of comparator selection in pharmacoepidemiologic research. Current epidemiology reports, 5:272–283.
\bibitem{b4}for Disease Control, C., (CDC, P., et al. (2008). Update on overall prevalence of major birth defects– atlanta, georgia, 1978-2005. MMWR. Morbidity and mortality weekly report, 57(1):1–5.
\bibitem{b5}Klein, A. Z., Cai, H., Weissenbacher, D., Levine, L. D., and Gonzalez-Hernandez, G. (2020). A natural language processing pipeline to advance the use of twitter data for digital epidemiology of adverse pregnancy outcomes. Journal of Biomedical Informatics, 112:100076.
\bibitem{b6}Steiner, J. M., Lokken, E., Bayley, E., Pechan, J., Curtin, A., Buber, J., and Albright, C. (2021). Cardiac and pregnancy outcomes of pregnant patients with congenital heart disease according to risk classification system. The American journal of cardiology, 161:95–101.
\bibitem{b7}Takami, M., Aoki, S., Kurasawa, K., Okuda, M., Takahashi, T., and Hirahara, F. (2014). A classifica- tion of congenital uterine anomalies predicting pregnancy outcomes. Acta Obstetricia et Gyneco- logica Scandinavica, 93(7):691–697.

\bibitem{b8} Bai, J., Wong, F. W., Bauman, A., \& Mohsin, M. (2002). Parity and pregnancy outcomes. American journal of obstetrics and gynecology, 186(2), 274-278.

\bibitem{b9} Pulley, L., Klerman, L. V., Tang, H., \& Baker, B. A. (2002). The extent of pregnancy mistiming and its association with maternal characteristics and behaviors and pregnancy outcomes. Perspectives on sexual and reproductive health, 206-211.


\bibitem{b10} Lean, S. C., Derricott, H., Jones, R. L., \& Heazell, A. E. (2017). Advanced maternal age and adverse pregnancy outcomes: A systematic review and meta-analysis. PloS one, 12(10), e0186287.

\bibitem{b11} Enomoto, K., Aoki, S., Toma, R., Fujiwara, K., Sakamaki, K., \& Hirahara, F. (2016). Pregnancy outcomes based on pre-pregnancy body mass index in Japanese women. PloS one, 11(6), e0157081.


\bibitem{b12} Shapiro, S., Jones, E. W., \& Densen, P. M. (1962). A life table of pregnancy terminations and correlates of fetal loss. The Milbank Memorial Fund Quarterly, 40(1), 7-45.

\bibitem{b13} Xu J, Murphy SL, Kochanek K, Bastian B, Arias E. Deaths: final data for 2016. Natl Vital Stat Rep. 2018;67(5).

\bibitem{b14} Ammon Avalos L, Galindo C, Li DK. A systematic review to calculate background miscarriage rates using life
table analysis. Birth Defects Res A Clin Mol Teratol. 2012;94(6):417-423.

\bibitem{b15} Rynn L, Cragan J, Correa A. Update on overall prevalence of major birth defects—Atlanta, Georgia, 1978-2005.
MMWR Morb Mortal Wkly Rep. 2008;57(1):1-5.


\bibitem{b16} MacDorman MF, Gregory ECW. Fetal and perinatal mortality: United States, 2013. Natl Vital Stat Rep.
2015;64(8).

\bibitem{b17} Ferré C, Callaghan W, Olson C, Sharma A, Barfield W. Effects of maternal age and age-specific preterm birth
rates on overall preterm birth rates—United States, 2007 and 2014. MMWR Morb Mortal Wkly Rep.
2016;65(43):1181-1184

\bibitem{b18} White, J., Fu, Q., Hays, S., Sandborn, M., Olea, C., Gilbert, H., Elnashar, A., Spencer-Smith, J. and Schmidt, D.C., 2023. A prompt pattern catalog to enhance prompt engineering with chatgpt. arXiv preprint arXiv:2302.11382.

\bibitem{b19} E.Ma,“Nlpaugmentation,”https://github.com/makcedward/nlpaug, 2019.

\bibitem{b20} Wu, Y., \& Hu, G. (2023, December). Exploring prompt engineering with GPT language models for document-level machine translation: Insights and findings. In Proceedings of the Eighth Conference on Machine Translation (pp. 166--169).

\bibitem{b21} Klein, A. Z., Sarker, A., Cai, H., Weissenbacher, D., \& Gonzalez-Hernandez, G. (2018). Social media mining for birth defects research: a rule-based, bootstrapping approach to collecting data for rare health-related events on Twitter. Journal of Biomedical Informatics, 87, 68--78.

\bibitem{b22} Sarker, A., Chandrashekar, P., Magge, A., Cai, H., Klein, A., \& Gonzalez, G. (2017). Discovering cohorts of pregnant women from social media for safety surveillance and analysis. Journal of Medical Internet Research, 19(10), e361.

\bibitem{b23} Klein, A. Z., \& Gonzalez-Hernandez, G. (2020). An annotated data set for identifying women reporting adverse pregnancy outcomes on Twitter. Data in Brief, 32, 106249.


\end{thebibliography}
\end{document}